\documentclass{article}

\usepackage{microtype}
\usepackage{graphicx}
\usepackage{subfigure}
\usepackage{booktabs} 
\usepackage{multirow}

\usepackage{hyperref}

\usepackage[accepted]{icml2025}

\usepackage{amsmath}
\usepackage{amssymb}
\usepackage{mathtools}
\usepackage{amsthm}
\usepackage{tabularx}

\usepackage[capitalize,noabbrev]{cleveref}

\theoremstyle{plain}

\theoremstyle{definition}

\theoremstyle{remark}

\usepackage[textsize=tiny]{todonotes}

\icmltitlerunning{Playmate: Flexible Control of Portrait Animation via 3D-Implicit Space Guided Diffusion}

\begin{document}

\twocolumn[
\icmltitle{Playmate: Flexible Control of Portrait Animation \\
            via 3D-Implicit Space Guided Diffusion}



\icmlsetsymbol{equal}{*}
\icmlsetsymbol{pl}{†}

\begin{icmlauthorlist}
\icmlauthor{Xingpei Ma}{equal}
\icmlauthor{Jiaran Cai}{equal,pl}
\icmlauthor{Yuansheng Guan}{equal}
\icmlauthor{Shenneng Huang}{}
\icmlauthor{Qiang Zhang}{}
\icmlauthor{Shunsi Zhang}{}
\end{icmlauthorlist}
{
\begin{center}
    \centering
    Guangzhou Quwan Network Technology 
    \vskip 0.05in
    \href{https://playmate111.github.io/Playmate/}{https://playmate.github.io/Playmate/}
\end{center}
}
{
\begin{center}
    \centering
    \vskip 0.1in
    \includegraphics[width=\linewidth]{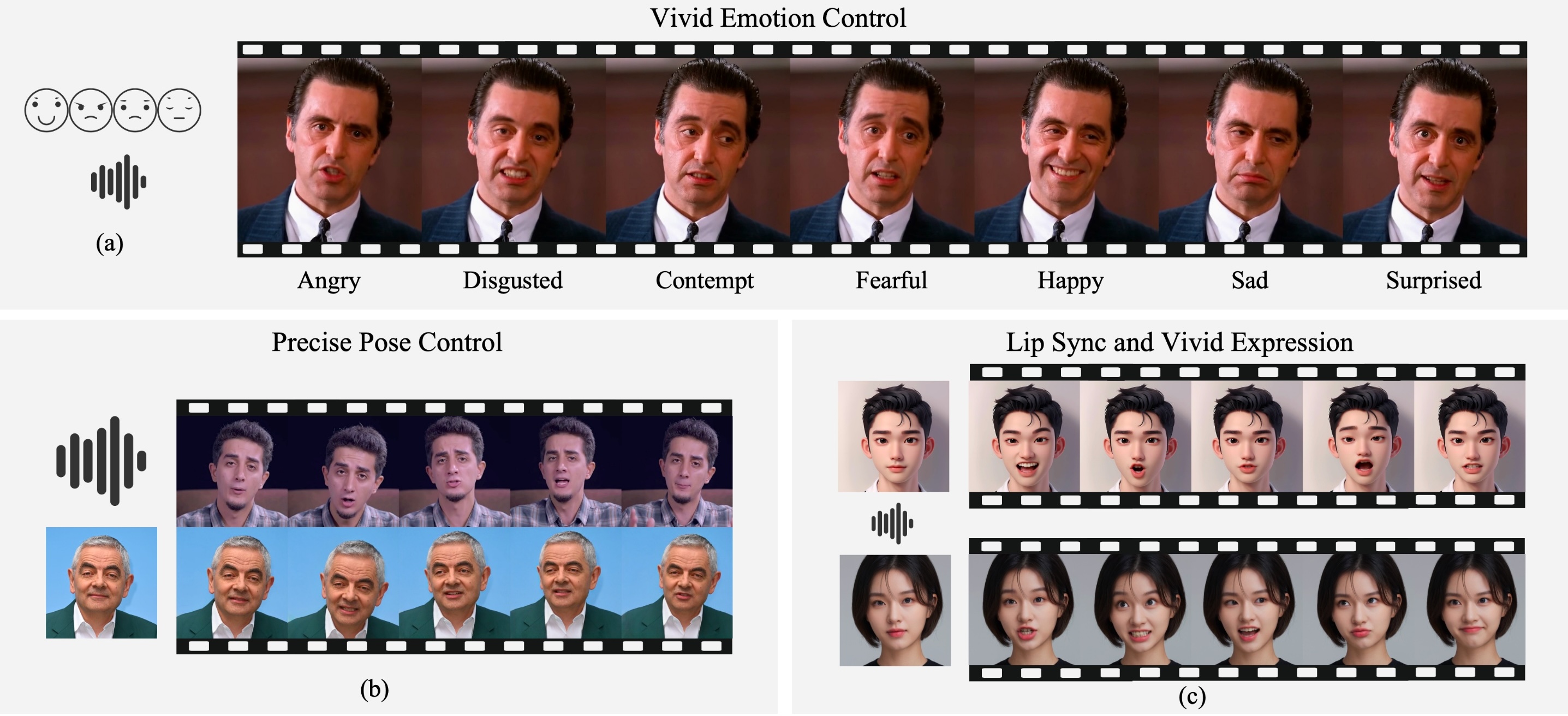}
    \begin{flushleft}
    {{\sl{Figure 1.}} Playmate can generate lifelike talking faces for arbitrary identity, guided by a speech audio clip and a variety of optional control conditions. (a) shows the generation results under different emotional conditions using the same audio clip. The top row in (b) shows the driving images, while the bottom row shows the generated results. The poses in the generated results are controlled by the driving images, and the lip movements are guided by the driving audio. (c) demonstrates highly accurate lip synchronization and vivid, rich expressions across different style images.}
    \end{flushleft}
\end{center}
}

\icmlcorrespondingauthor{Jiaran Cai}{caijiaran@52tt.com}
\icmlcorrespondingauthor{Xingpei Ma}{maxingpei@52tt.com}

\icmlkeywords{Diffusion, Portrait Animation, Transformer, Audio-driven}

\vskip 0.3in
]



\printAffiliationsAndNotice{\icmlEqualContribution\icmlProjectLead} 

\begin{abstract}
Recent diffusion-based talking face generation models have demonstrated impressive potential in synthesizing videos that accurately match a speech audio clip with a given reference identity. However, existing approaches still encounter significant challenges due to uncontrollable factors, such as inaccurate lip-sync, inappropriate head posture and the lack of fine-grained control over facial expressions. 
In order to introduce more face-guided conditions beyond speech audio clips, a novel two-stage training framework \textbf{\textit{Playmate}} is proposed to generate more lifelike facial expressions and talking faces. 
In the first stage, we introduce a decoupled implicit 3D representation along with a meticulously designed \textbf{\textit{motion-decoupled module}} to facilitate more accurate attribute disentanglement and generate expressive talking videos directly from audio cues. Then, in the second stage, we introduce an \textbf{\textit{emotion-control module}} to encode emotion control information into the latent space, enabling fine-grained control over emotions and thereby achieving the ability to generate talking videos with desired emotion. Extensive experiments demonstrate that Playmate not only outperforms existing state-of-the-art methods in terms of video quality, but also exhibits strong competitiveness in lip synchronization while offering improved flexibility in controlling emotion and head pose. The code will be available at \href{https://github.com/Playmate111/Playmate}{https://github.com/Playmate111/Playmate}. 
\end{abstract}

\section{Introduction}
\label{introduction}
Audio-driven portrait animation techniques \cite{jiang2024audio,xue2024human} aim to synthesize a lifelike talking face from a static image and speech audio, thereby creating realistic and expressive avatars. In recent years, with the significant development of diffusion-based models \cite{ho2020denoising,song2020denoising,Dhariwal_Nichol_2021,rombach2022high}, this field has attracted increasing attention from both academia and industry. Consequently, these advances have unlocked substantial potential for diverse applications, including film dubbing, video generation, and interactive media.

Existing audio-driven portrait animation approaches can be broadly categorized into two main branches: GAN/NeRF-based methods \cite{chen2018lip,prajwal2020lip,lu2021live,zhang2023dinet,ma2023styletalk,tan2024flowvqtalker,ki2023stylelipsync,cheng2022videoretalking,guo2021ad,wang2024expression,ye2023geneface++} and diffusion-based methods \cite{shen2023difftalk,sun2023vividtalk,tian2025emo,he2023gaia,xu2024vasa,xu2024hallo,zheng2024memo,cao2024joyvasa,ji2024sonic,ma2023dreamtalk,chen2024echomimic,lin2024takin,liu2024anitalker,sun2024diffposetalk}. The former category has primarily focused on the accuracy of lip synchronization \cite{chung2017out} and has delivered significant results in this regard. However, GAN/NeRF-based methods often overlook the holistic coordination between audio cues and facial expressions, leading to a failure in generating expressive facial dynamics and lifelike expressions. Recently, the advent of diffusion models has facilitated the generation of high-quality images and videos \cite{podell2023sdxl,esser2024scaling,liu2024sora,lin2024open,kong2024hunyuanvideo,hong2022cogvideo,yang2024cogvideox}. Several studies have introduced diffusion models into the field of portrait animation, enabling them to excel in generating talking face videos. Nonetheless, the expression, lip-sync, and head pose of videos generated by diffusion-based methods are strongly correlated and coupled with audio cues, which limits flexibility in controlling specific facial attributes such as head pose and emotional expression, making it challenging to modify one attribute independently without altering the associated audio content. This dependency limits the customization and adaptability of generated animations for different applications. 

Based on the above observations, we propose Playmate, a novel two-stage training framework that leverages a 3D-Implicit Space Guided Diffusion Model to generate lifelike talking faces with controllable facial attributes. To achieve this goal, we first introduce a decoupled implicit 3D representation proposed in face-vid2vid \cite{wang2021one} and LivePortrait \cite{guo2024liveportrait}. This implicit representation effectively disentangles multiple face attributes, including expression, lip movement, head pose, and others, thereby enabling more flexible editing and control of these attributes. Subsequently, we train an audio-conditioned diffusion transformer combined with a carefully designed \textbf{\textit{motion-decoupled module}}, facilitate more accurate attribute disentanglement and generate expressive talking videos directly from audio cues. Finally, to enhance the controllability of emotions, we introduce the \textbf{\textit{emotion-control module}}, which utilizes DiT blocks \cite{peebles2023scalable} to encode specified emotion conditions and integrates the encoded information into the aforementioned audio-conditioned diffusion model, thereby achieving flexible control over emotions and improving the editability of the generated talking faces. As shown in Figure 1, Playmate has advantages in audio-driven portrait animation. The main contributions of this paper can be summarized as follows: 

\begin{itemize}
\item We present Playmate, a novel framework that utilizes 3D-Implicit Space Guided Diffusion for generating talking face videos. 
\item We meticulously designed a motion-decoupled module and trained a diffusion transformer to improve motion disentanglement and generate motion sequences directly from audio cues. 
\item A novel emotion-control module and corresponding training strategy are proposed to encode emotion information into latent space and enhance the controllability of emotions of talking head video. 
\item Experiments show that our method achieves SOTA performance in terms of video quality, motion diversity and emotion controllability. 
\end{itemize}

\section{Related Work}

\subsection{GAN/NeRF-based Audio-driven Portrait Animation}
Talking face video generation has been a long-standing challenge in computer vision and graphics. The goal is to synthesize lifelike and synchronized talking videos from driving audio and static reference images. Early GAN/NeRF-based approaches, such as Wav2Lip \cite{prajwal2020lip}, Dinet \cite{zhang2023dinet}, and VideoReTalking \cite{cheng2022videoretalking}, primarily focused on achieving high-quality lip-sync while keeping other facial attributes static. Consequently, these methods fail to capture strong correlations between the audio and other facial attributes, such as facial expression and head movements. To address this limitation, GANimation \cite{pumarola2018ganimation} introduced an unsupervised method to generate talking videos with a specific expression. EAMM \cite{ji2022eamm} synthesizes emotional talking faces with augmented emotional source videos. More recent studies have typically employed intermediate motion representations (e.g., landmark coordinates, 3D facial mesh, and 3DMM) to generate videos from audio \cite{zhang2023sadtalker,gan2023efficient,peng2024synctalk}. However, such approaches often generate inaccurate intermediate representations, which restricts the expressiveness and realism of the resulting videos. In contrast, our framework generates accurate motion representations based on diffusion transformer. 

\subsection{Diffusion-based Audio-driven Portrait Animation}
Diffusion models have shown impressive performance across various vision tasks. However, previous \cite{bigioi2024speech,mukhopadhyay2024diff2lip,shen2023difftalk} attempts to utilize diffusion models for generating talking heads have only yielded neutral-emotion expressions, leading to unsatisfactory results. Some of the latest methods have some optimizations for this purpose, such as EMO \cite{tian2025emo}, Hallo \cite{xu2024hallo}, Echomimic \cite{chen2024echomimic}, and Loopy \cite{jiang2024loopy}. EMO introduces a novel framework that ensures consistency in audio-driven animations across video frames, thereby enhancing the stability and naturalness of synthesized speech animations. Hallo contributes a hierarchical audio-driven method for animating portrait images, tackling the complexities of lip synchronization, expression, and pose alignment. MEMO \cite{zheng2024memo} proposes an end-to-end audio-driven portrait animation approach to generate identity-consistent and expressive talking videos. Several of the above methods can generate vivid portrait videos by fine-tuning pre-trained diffusion models. However, they usually use coupled latent spaces to represent facial attributes in relation to the audio. Facial attributes such as expression and head posture are often generated directly from audio cues. This coupling limits the ability to customize control over certain facial attributes, such as pose and expression. In Playmate, we leverage a 3D implicit space that decoupled various facial attributes, enabling diverse and controllable facial animations while maintaining high accuracy in lip synchronization. 

\subsection{Facial Representation in Audio-driven Portrait Animation}
Facial representation learning has been extensively studied in previous works. Various methods \cite{siarohin2019first,ren2021pirenderer,li2017learning} disentangled variables using 3DMM, sparse keypoints, or FLAME to explicitly characterize facial attributes. Furthermore, in the field of audio-driven portrait animation, several studies have introduced facial representation techniques to generate lifelike talking videos. Sadtalker \cite{zhang2023sadtalker} separates generation targets into different categories, including eye blinks, head poses, and lip-only 3DMM coefficients. Recent works such as VASA-1 \cite{xu2024vasa}, Takin-ADA \cite{lin2024takin}, DreamTalk \cite{ma2023dreamtalk}, and JoyVASA \cite{cao2024joyvasa} have begun to combine face representations with diffusion models to achieve more naturalistic results. Inspired by these advancements, we similarly utilize face representation techniques to generate more natural and controllable talking videos. 
\setcounter{figure}{1}
\begin{figure*}[!t]
    \centering
    \includegraphics[width=0.80\linewidth]{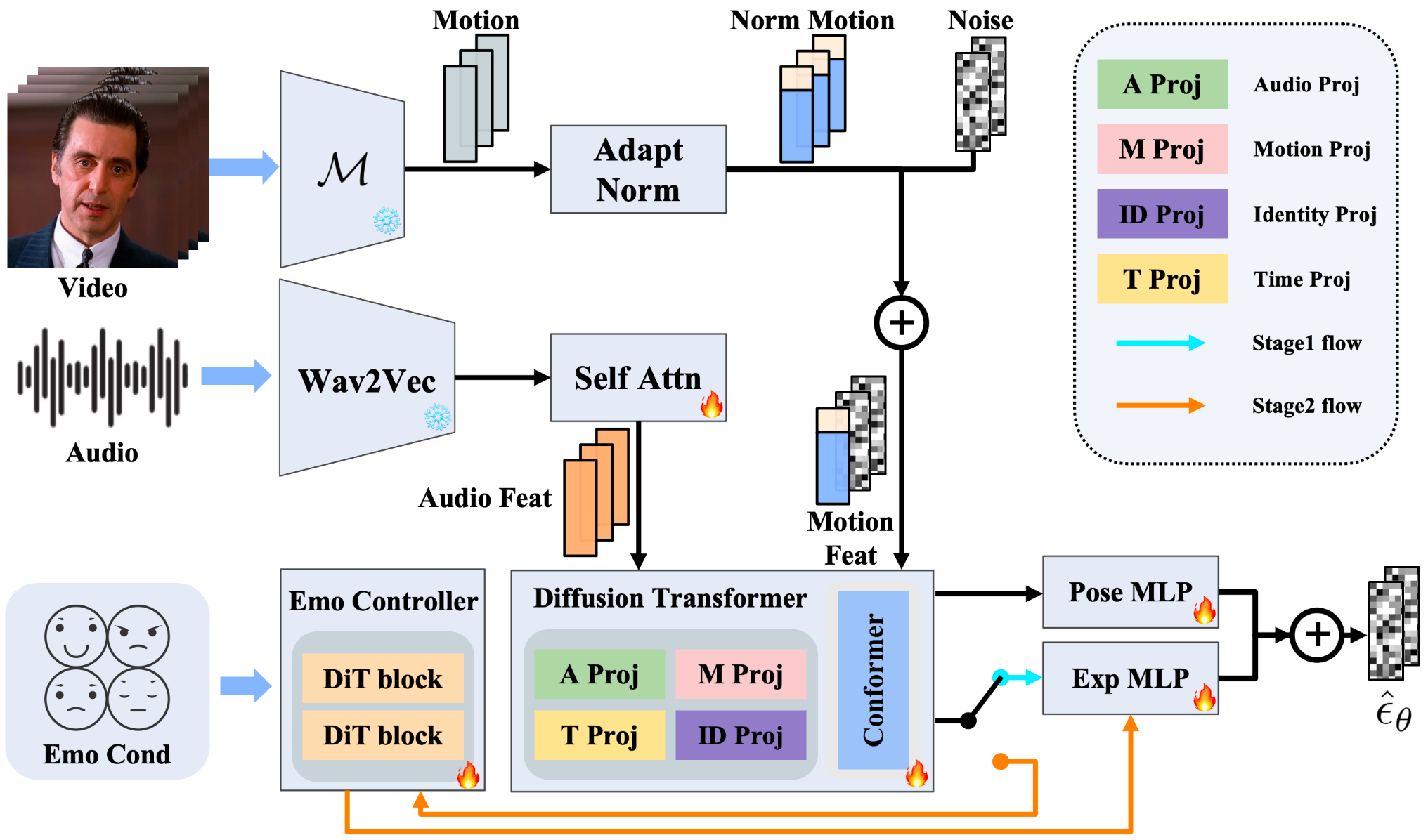}
    \caption{\textbf{Framework of our approach.} Playmate is a two-stage training framework that leverages a 3D-Implicit Space Guided Diffusion Model to generate lifelike talking faces. In the first stage, Playmate utilizes a motion-decoupled module to enhance attribute disentanglement accuracy and trains a diffusion transformer to generate motion sequences directly from audio cues. In the second stage, we use an emotion-control module to encode emotion control information into the latent space, enabling fine-grained control over emotions, thereby improving flexibility in controlling emotion and head pose.}
    \label{fig:framework}
    \vskip -0.1in
\end{figure*}

\section{Methodology}
As shown in \cref{fig:framework}, \textbf{\textit{Playmate}} uses a 3D implicit space as the intermediate representation for generating talking heads from a single static image, guided by a speech audio clip and a set of optional control signals. This section elaborates on our method in detail. We begin with a brief introduction of the 3D implicit space. Furthermore, we describe our meticulously designed approach for generating motion sequences directly from audio cues. Finally, we introduce our emotion-control module and two-stage training strategy, which enhance the controllability of emotions of talking head video. 

\subsection{Expressive and Disentangled Latent Face Space Construction}
\label{subsec:latent_space}

Facial representation aims to construct a latent face space that exhibits high degrees of expressiveness and disentanglement. Typically, this approach separates various aspects of facial data into distinct components, such as appearance features and motion attributes. In the field of audio-driven portrait animation, many studies have leveraged these latent spaces to generate talking heads. For example, VASA-1 based its model on the 3D-aid face reenactment framework from \cite{wang2021one,drobyshev2022megaportraits}, while Takin-ADA and JoyVASA constructed their latent space based on face-vid2vid and LivePortrait, respectively. Similarly, we adopt and enhance the decoupled facial representation proposed by face-vid2vid and LivePortrait. 

LivePortrait primarily comprises an appearance feature extractor $\mathcal{F}$, a motion extractor $\mathcal{M}$, a warping module $\mathcal{W}$, and a decoder $\mathcal{G}$.
When presented with a source image $I_s$ and a driving image $I_d$, LivePortrait initially utilizes $\mathcal{F}$ to extract appearance feature $f_s$ from $I_s$, and separately captures the motion information from both $I_s$ and $I_d$ using $\mathcal{M}$. 
Subsequently, it leverages the motion information extracted from $I_d$ to animate $I_s$ using $\mathcal{W}$ and $\mathcal{G}$, ensuring that the animated result retains the original appearance while adopting the motion cues from the driving image. 
The motion information is characterized by canonical keypoints $x_c\in\mathbb{R}^{K\times3}$, expression deformations $\delta\in\mathbb{R}^{K\times3}$, a rotation matrix $R\in\mathbb{R}^{3\times3}$, a translation vector $t\in\mathbb{R}^{3}$, and a scaling factor $s\in\mathbb{R}$. The transformation in LivePortrait is formalized as: 
\begin{equation}
    \left\{
        \begin{array}{cc}
            x_s=s_s\cdot(x_{c,s}R_s+\delta_s)+t_s,\\
            x_d=s_d\cdot(x_{c,s}R_d+\delta_d)+t_d,
        \end{array}
    \right.
\end{equation}
where $x_{c,s}$ represents the canonical keypoints of the source image $I_s$, and $x_s$ and $x_d$ are the source and driving 3D implicit keypoints after transformation. Additional information about this framework can be found in reference \cite{guo2024liveportrait}.

Similar to VASA-1, we found that the original LivePortrait also suffered from poor disentanglement between facial dynamics and head pose. To address this issue, we utilize the pairwise head pose and facial dynamics transfer loss used in VASA-1 to improve its disentanglement.
Specifically, let $I_i$ and $I_j$ be two randomly sampled frames from the same video clip. Then we transfer the head pose of $I_j$ onto $I_i$, resulting in $\hat{I}_{i,j^{pose}}=\mathcal{G}(\mathcal{W}(f_i,x_i,x_{i,j^{pose}}))$, the transformation is formalized as:
\begin{equation}
    \left\{
        \begin{array}{ll}
            x_i=s_i\cdot(x_{c,i}R_i+\delta_i)+t_i,\\
            x_{i,j^{pose}}=s_j\cdot(x_{c,i}R_j+\delta_i)+t_j.
        \end{array}
    \right.
\end{equation}
Similarly, we can transfer $I_i$'s expression onto $I_j$, yielding $\hat{I}_{j,i^{exp}}=\mathcal{G}(\mathcal{W}(f_j,x_j,x_{j,i^{exp}}))$, and the transformation is formalized as: 
\begin{equation}
    \left\{
        \begin{array}{ll}
            x_j=s_j\cdot(x_{c,j}R_j+\delta_j)+t_j,\\
            x_{j,i^{exp}}=s_j\cdot(x_{c,j}R_j+\delta_i)+t_j.
        \end{array}
    \right.
\end{equation}
Ultimately, a perceptual loss \cite{johnson2016perceptual} is employed between $\hat{I}_{i,j^{pose}}$ and $\hat{I}_{j,i^{exp}}$ to enhance the disentanglement between facial dynamics and head pose, the loss function as:
\begin{equation}
\mathcal{L}_p=\parallel \mathcal{V}(\hat{I}_{i,j^{pose}}) - \mathcal{V}(\hat{I}_{j,i^{exp}})\parallel_2,
\end{equation}
where $\mathcal{V}$ denotes the feature extractor of VGG19 \cite{simonyan2014very}.

\subsection{Audio Conditioned Diffusion Transformer}
After constructing the latent face space, we can extract motions by utilizing the frozen $\mathcal{M}$ and train the motion generator, with audio cues serving as the condition.

\textbf{Adaptive Normalization.} To achieve a more effective decoupling of expression and head pose, we employ adaptive normalization by using different means and standard deviations when training the motion generator. Specifically, for expression, we compute the global mean and standard deviation using the entire training dataset, as: 
\begin{equation}
    \left\{
        \begin{array}{ll}
            \mu^{\delta}={{\sum_{i=1}^{M}{\sum_{j=1}^{N_i}{\delta_{i,j}}}}\over\sum_{i=1}^{M}{N_i}} ,\\
            \sigma^{\delta}=\sqrt{\sum_{i=1}^{M}{\sum_{j=1}^{N_i}{(\delta_{i,j}-\mu^{\delta})^2}}\over\sum_{i=1}^{M}{N_i}},
        \end{array}
    \right.
\end{equation}
where $N_i$ is the number of frames in the $i$-th video clip, and $M$ is the total number of all training samples.
For head pose, we treat each identity independently and compute the private mean and standard deviation for each sample along the time dimension, as:
\begin{equation}
    \left\{
        \begin{array}{ll}
            \mu^{\rho}_{i}={\sum_{j=1}^{N_i}{\rho_{i,j}}} ,\\
            \sigma^{\rho}_{i}=\sqrt{\sum_{j=1}^{N_i}{(\rho_{i,j}-\mu^{\rho}_{i})^2}\over{N_i}},
        \end{array}
    \right.
\end{equation}
where $\rho$ denotes the head pose information.
By combining adaptive normalization with the transfer loss mentioned in \ref{subsec:latent_space}, we achieve better decoupling of motion. We refer to this approach as the \textbf{motion-decoupled module}.

\textbf{Speech Representation.} Extensive research has demonstrated that pre-trained speech models, such as Wav2Vec2 \cite{baevski2020wav2vec} and HuBERT \cite{hsu2021hubert}, outperform traditional features like MFCC in performance. Similar to other methods, we utilize Wav2Vec2 as our speech encoder to extract audio features. Additionally, we use a self-attention module to align audio features with motion features. 

\textbf{Diffusion Transformer.} The architecture of the Diffusion Transformer is illustrated in \cref{fig:framework}. We employ four Proj modules, primarily composed of fully connected layers, to extract semantic information from four unique features, aligning this information across different modalities. Next, we utilize a multilayer Conformer \cite{gulati2020conformer}, a Pose-MLP, and a Exp-MLP in combination with the diffusion formulation to generate motion sequences. Diffusion models define two Markov chains: the forward chain progressively adds Gaussian noise to the target data, while the reverse chain iteratively refines the raw signal from the noise. During training, we gradually transform clean motion $m$ into Gaussian noise $m_t$ following the principles of Denoising Diffusion Probabilistic Models(DDPM) \cite{ho2020denoising}. The Diffusion Transformer is then trained to reverse this noise-adding process by taking $m_t$ and other conditional features as input and predicting the added noise $\epsilon$. The objective function for training can be expressed as: 
\begin{equation}
{\mathcal{L}_{diff}}=\mathbb{E}_{m_t,f_a,f_{id},t,\epsilon}\left(\left\|\epsilon-\hat{\epsilon}_\theta \left(m_t,f_a,f_{id},t\right)\right\|^{2}\right),
\end{equation}
where $f_a$, $f_{id}$ are the audio feature and identity feature, $\hat{\epsilon}_\theta$ represents the noise prediction made by the Diffusion Transformer.

\subsection{Emotion-control Module}
After completing the first training stage, we obtain a diffusion transformer that generates motion sequences guided by audio cues. To enhance the controllability of emotions in talking head videos, we propose the emotion-control module. This involves fixing the parameters of the diffusion transformer and training an emotion controller based on the DiT block \cite{peebles2023scalable}. 
\begin{figure}[t]
    \centering
    \includegraphics[width=0.8\linewidth]{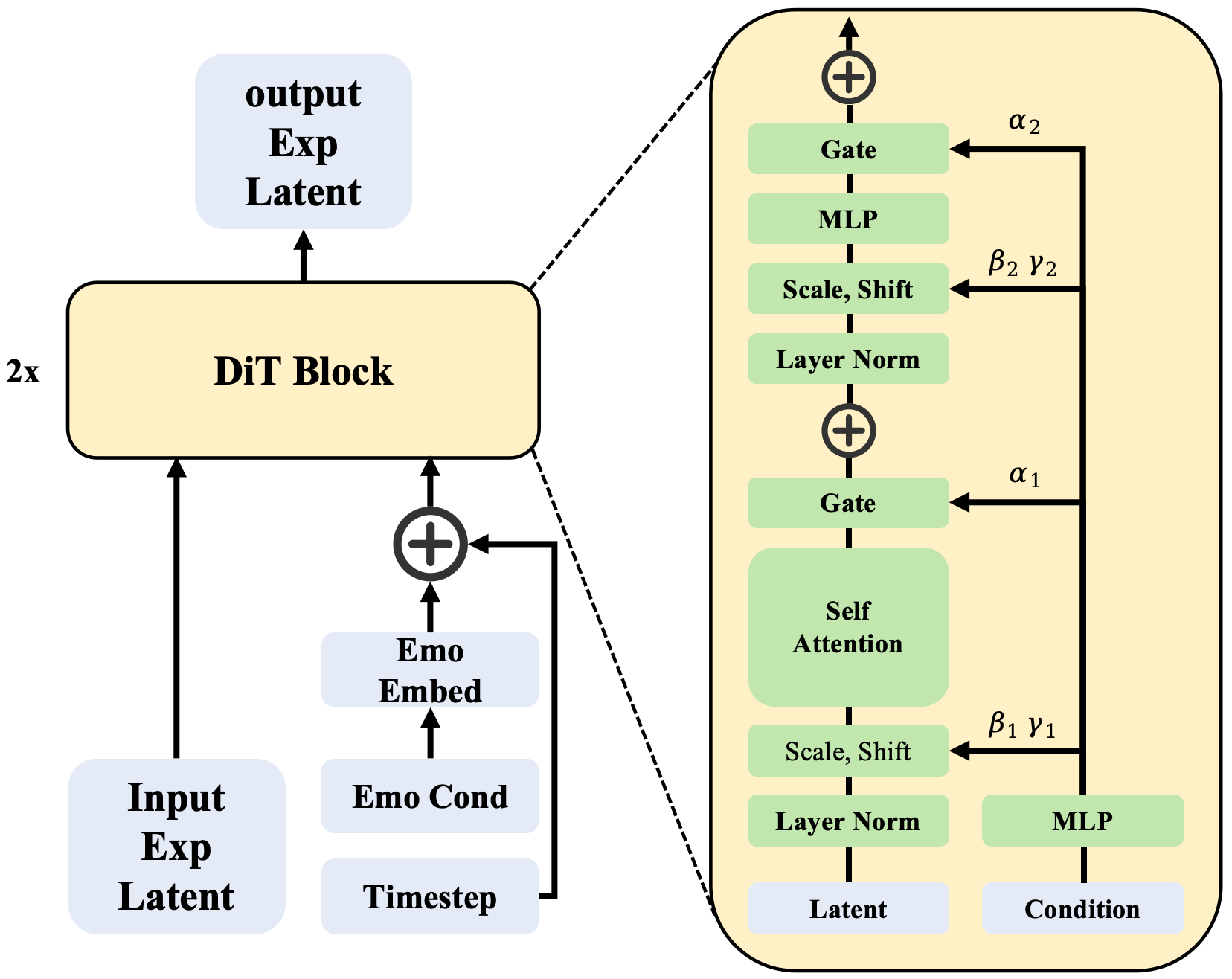}
    \caption{The structure of the Emotion-control Module.}
    \vspace{-15pt}
    \label{fig:emo_controller}
\end{figure}
As shown in \cref{fig:framework} and \cref{fig:emo_controller}, the emotion-control module consists of two DiT blocks. The first block receives inputs from the conformer and the emotion condition, while the second block takes the output of the first block and emotion condition as its inputs. The outputs of the second block are then passed to Exp-MLP, replacing the original conformer's output. Through the aforementioned operations, we encode emotion control information into the features and pass them to the Exp-MLP module, thereby achieving the ability to generate talking videos with desired emotion. 

During the second phase of training, we freeze the parameters of the diffusion transformer and train only the parameters of the emotion-control module. Experimental results demonstrate that this approach effectively maintains lip-sync accuracy while integrating emotion conditions, allowing for precise emotion control in the generated animations.

\textbf{Classifier-free guidance(CFG).} We adopt a classifier-free guidance \cite{ho2022classifier} approach following \cite{brooks2023instructpix2pix}, which has been successfully applied to image generation from multiple conditions. During training, a random dropout strategy is applied to each of the input conditions to improve the model's robustness and generalization. During inference, we apply:
\begin{equation}
\begin{split}
\hat{\epsilon}_\theta=&\hat{\epsilon}_\theta \left(m_t,\emptyset,\emptyset,f_{id},t\right)\\&+w_a[\hat{\epsilon}_\theta \left(m_t,f_{a},\emptyset,f_{id},t\right)-\hat{\epsilon}_\theta \left(m_t,\emptyset,\emptyset,f_{id},t\right)]\\&+w_{e}[\hat{\epsilon}_\theta \left(m_t,f_a,f_{e},f_{id},t\right)-\hat{\epsilon}_\theta \left(m_t,f_a,\emptyset,f_{id},t\right)],
\end{split}
\label{eq:cfg}
\end{equation}
where $w_a$ and $w_{e}$ are the guidance scales for audio condition and emotion condition, respectively. 

\begin{table*}[!t]
  \vskip -0.1in
  \caption{\textbf{Quantitative comparisons of video quality and lip synchronization with state-of-the-art methods on two test datasets.} The best results are in \textbf{bold}, and the second-best are in \underline{underlined}. Playmate consistently outperforms existing methods in terms of video quality and identity preservation, while also exhibiting strong competitiveness in lip synchronization.}
  \vskip 0.1in
  \centering
  \scalebox{0.9}{
  \begin{tabular}{c|c|cccccc}
  \toprule
  Dataset & Method & FID $\downarrow$ & FVD $\downarrow$ & Sync-C $\uparrow$ & Sync-D $\downarrow$ & CSIM $\uparrow$ & LPIPS $\downarrow$ \\\hline
   \multirow{6}*{HDTF} & Hallo \cite{xu2024hallo} & 30.484 & 288.479 & 7.923 & 7.531 & 0.804 & 0.139 \\
   ~ & Hallo2 \cite{cui2024hallo2} & 30.768 & \underline{288.385} & 7.754 & 7.649 & 0.822 & 0.138 \\
   ~ & MEMO \cite{zheng2024memo} & \underline{27.713} & 299.493 & 8.059 & 7.473 & \underline{0.840} & \underline{0.132} \\
   ~ & Sonic \cite{ji2024sonic} & 29.189 & 305.867 & \textbf{9.139} & \textbf{6.549} & 0.783 & 0.149 \\
   ~ & JoyVASA \cite{cao2024joyvasa} & 29.581 & 306.683 & 8.522 & 7.215 & 0.781 & 0.157 \\
   ~ & Playmate(Ours) & \textbf{19.138} & \textbf{231.048} & \underline{8.580} & \underline{6.985} & \textbf{0.848} & \textbf{0.099} \\
   \hline
    \multirow{6}*{Collected dataset} & Hallo \cite{xu2024hallo} & 46.114 & 288.415 & 6.454 & 8.384 & 0.767 & 0.139 \\
   ~ & Hallo2 \cite{cui2024hallo2} & 46.185 & 295.532 & 6.509 & 8.358 & 0.761 & 0.144 \\
   ~ & MEMO \cite{zheng2024memo} & 39.224 & 260.498 & 6.569 & 8.193 & \underline{0.782} & \underline{0.130} \\
   ~ & Sonic \cite{ji2024sonic} & \underline{39.069} & \underline{254.959} & \textbf{7.972} & \textbf{7.124} & 0.762 & 0.139 \\
   ~ & JoyVASA \cite{cao2024joyvasa} & 50.314 & 304.621 & 6.858 & 8.194 & 0.713 & 0.163 \\
   ~ & Playmate(Ours) & \textbf{34.716} & \textbf{227.871} & \underline{7.125} & \underline{8.007} & \textbf{0.797} & \textbf{0.128} \\
  \bottomrule
  \end{tabular}%
  }
  \label{tab:quantitative_comparison}%
\end{table*}

\begin{figure*}[!t]
    \centering
    \includegraphics[width=0.9\linewidth]{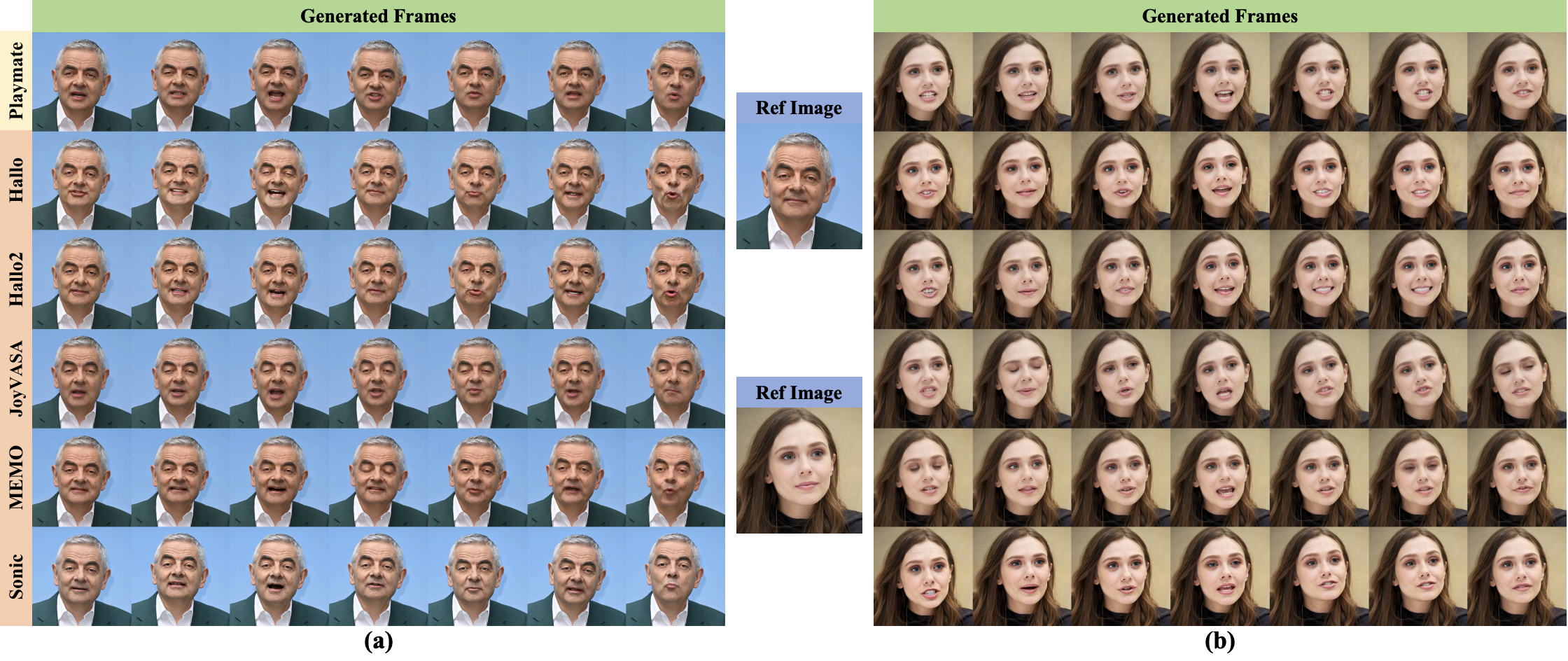}
    \vskip -0.15in
    \caption{\textbf{Qualitative comparisons with state-of-the-art methods.} Previous methods were prone to generate artifacts in tooth rendering(e.g., (a)-row 6, column 3; (b)-row 3, column 1) and lip synchronization(e.g., (a)-row 4, column 7; (b)-row 2, column 7). Conversely, our approach boasts a superior decoupling capability, which allows it to create more lifelike talking head videos. For more comparison details, please see the Appendix. }
    \label{fig:qualitative_comparison}
    \vskip -0.1in
\end{figure*}

\section{Experiments}

\subsection{Experiments Setup}

\textbf{Datasets.} We utilize a mixture of datasets, including AVSpeech \cite{ephrat2018looking}, CelebV-Text \cite{yu2023celebv}, Acappella \cite{montesinos2021cappella}, MEAD \cite{wang2020mead}, MAFW \cite{liu2022mafw}, and a talking video dataset collected by us to train our Playmate. In the first stage, we selected approximately 80,000 video clips from the AVSpeech, CelebV-Text, Acappella, and our own dataset to train the diffusion transformer. For the second phase, we selected approximately 30,000 emotionally labeled video clips from the MEAD, MAFW, and our own dataset to train the emotion control module. The duration of each training video ranges from 3 to 30 seconds. We set aside a portion of video clips from our dataset that were not involved in the training process as our out-of-distribution test set. Besides, we choose HDTF \cite{zhang2021flow} as our out-of-domain dataset , which comprises about 362 different videos with original resolution of 720P or 1080P. 

\textbf{Implementation Details.} In our experiments, the videos are initially converted to 25 fps and subsequently cropped to a resolution of 256 × 256 pixels based on face landmarks extracted using InsightFace \cite{deng2020retinaface,guo2021sample}. The final output resolution is set to 512 × 512 pixels. During preprocessing, the audios were resampled to 16kHz. The first training phase utilized four NVIDIA A100 GPUs over a 3-day period, with models initialized from scratch. In the second phase, we continued training for two days with two NVIDIA A100 GPUs, while freezing the parameters of the diffusion transformer. For all experiments, we employed the Adam optimizer \cite{kingma2014adam}. In the inference phase, multi-condition CFG is performed. The CFG scales of the audio $w_a$ and the emotion condition $w_e$ are set to 1.5.

\textbf{Evaluation Metrics.} We demonstrate the superiority of our method using multiple widely recognized metrics from previous studies. Specifically, we employ Fréchet Inception Distance (FID) \cite{heusel2017gans} and Fréchet Video Distance (FVD) \cite{unterthiner2019fvd} to evaluate the quality of the generated data. For evaluating lip synchronization and motion fluidity, we compute the confidence score (Sync-C) and feature distance (Sync-D) using the pretrained SyncNet \cite{chung2017out}. Additionally, we compute the cosine similarity (CSIM) of identity vectors extracted using the ArcFace \cite{deng2019arcface} face recognition model to evaluate identity preservation. Furthermore, we leverage Learned Perceptual Image Patch Similarity (LPIPS) \cite{zhang2018unreasonable} to measure the feature-level similarity between the generated faces and their ground-truth faces. For the evaluation of emotion control, we employ an additional emotion classifier from \cite{savchenko2023facial} to calculate the emotion accuracy (Emo-A) of the generated videos.

\begin{figure}[t]
    \centering
    \includegraphics[width=1.0\linewidth]{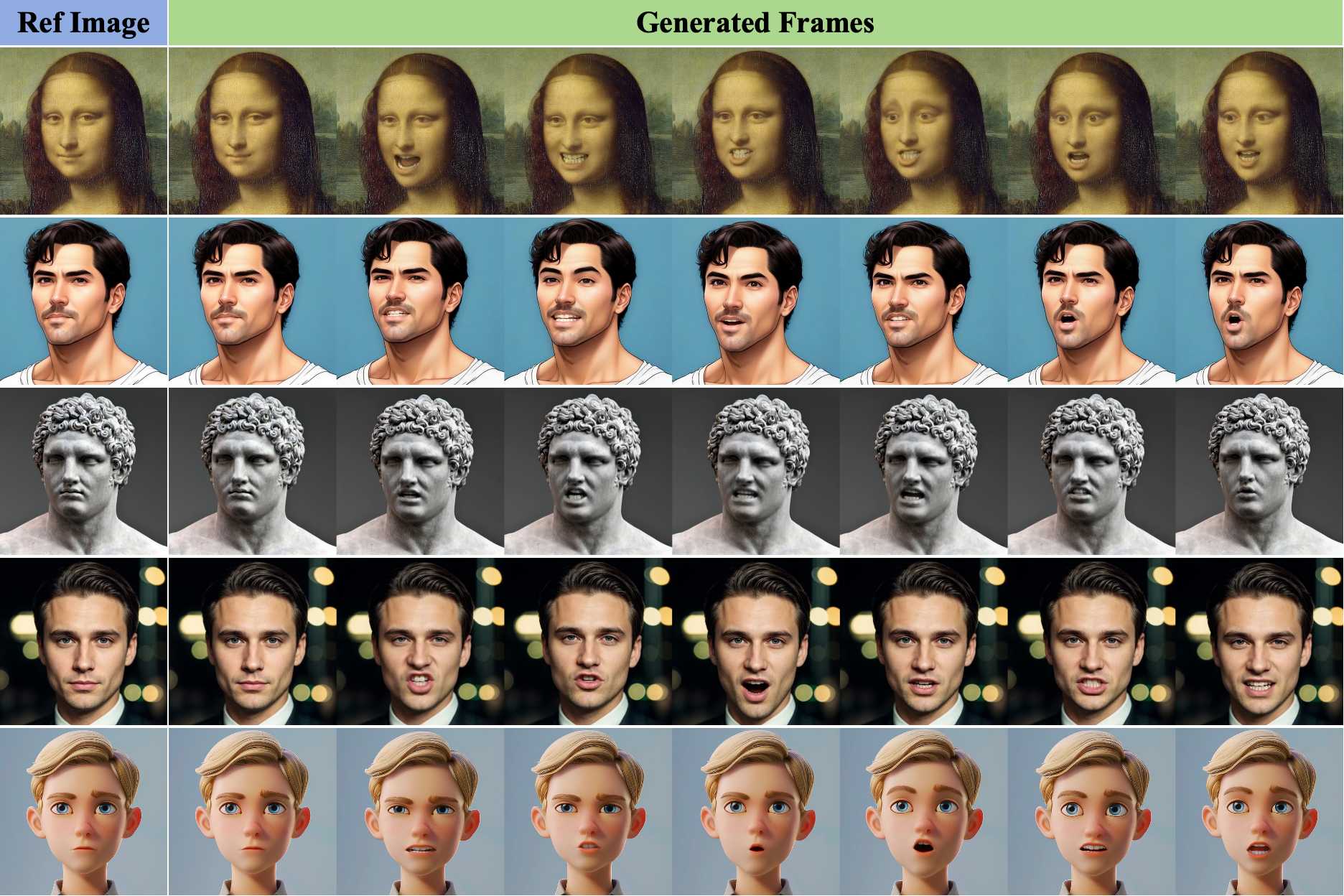}
    \vskip -0.1in
    \caption{\textbf{Visualization results in different style images.} Playmate can drive a wide range of portraits, including real humans, animations, artistic portraits, and even animals.}
    \label{fig:multi_style}
    \vskip -0.1in
\end{figure}

\subsection{Results and Analysis}
\textbf{Quantitative Results.} We benchmark our method against SOTA audio-driven portrait animation methods, including Hallo \cite{xu2024hallo}, Hallo2 \cite{cui2024hallo2}, JoyVASA \cite{cao2024joyvasa}, MEMO \cite{zheng2024memo}, and Sonic \cite{ji2024sonic}. As shown in \cref{tab:quantitative_comparison}, our Playmate significantly outperforms other methods in terms of FID, FVD, CSIM, and LPIPS on two test datasets, while also exhibiting strong competitiveness in lip synchronization. Regarding video quality, our method achieves the lowest FID and FVD scores on both test sets. Specifically, on the HDTF dataset, our FID and FVD scores are 30\% and 20\% lower than those of the second-best method, respectively, indicating superior video quality compared to other methods. For the CSIM and LPIPS metrics, we also achieve the best results, indicating superior performance in identity preservation and image quality. Additionally, our method achieves good results in Sync-C and Sync-D, both of which are second-best, exhibiting strong competitiveness. Although our quantitative metrics for lip synchronization on the two test sets are marginally lower than Sonic's, they still outperform other compared methods. Furthermore, Sonic is a purely audio-driven algorithm that generates all features of the portrait, including lip movements, expressions, pose, etc., based on audio. This indicates that its driving flexibility is limited. In contrast, we achieve multiple controllable portrait driving methods by constructing a precise attribute disentanglement space, offering users various flexible driving options. The implementation complexity of this decoupling and subsequent driving approach is higher than that of a simple audio-driven method.
\begin{figure}[t]
    \centering
    \includegraphics[width=1.0\linewidth]{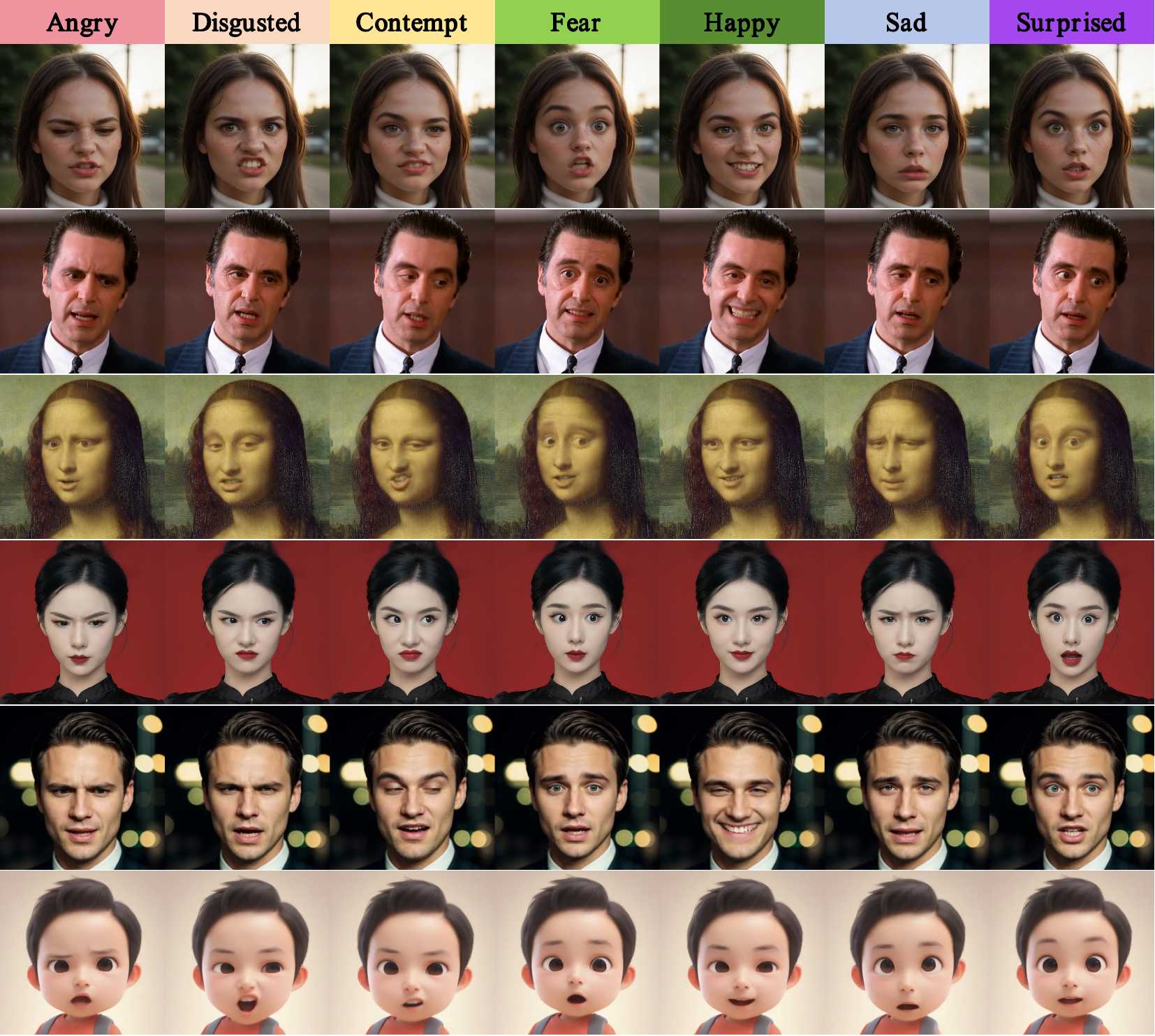}
    \vspace{-15pt}
    \caption{\textbf{Visualization results of emotion control.} Each row shows the generation for different identity under different emotional conditions using the same audio clip, demonstrating the flexibility in controlling emotion of Playmate.}
    \vskip -0.1in
    \label{fig:emo_control}
\end{figure}

\textbf{Qualitative Results.} \cref{fig:qualitative_comparison} provides a visualization comparison using open datasets. Upon analysis, previous methods were prone to generate artifacts in tooth rendering(e.g., (a)-row 6, column 3; (b)-row 3, column 1) and lip synchronization(e.g., (a)-row 4, column 7; (b)-row 2, column 7). Conversely, our approach boasts a superior decoupling capability, which allows it to create more lifelike talking head videos. For more comparison details, please see the Appendix. Since images cannot adequately reflect important aspects such as synchronization, naturalness, and stability, the full video comparison will be included in the supplementary materials or on our \href{https://playmate111.github.io/Playmate/}{project page}.

\textbf{Visualization Results in Different Style Images.} We further investigate the generation performance of our method in different style images. Given that our Playmate can directly generate motion sequences from audio, it is capable of driving a wide range of portraits, including real humans, animations, artistic portraits, and even animals. Although this versatility may lead to characters having similar expressions, by integrating the emotion-control module and adjusting the CFG weight, we can easily achieve diverse expressions and movements. As illustrated in \cref{fig:multi_style}, our method successfully generates a wide range of expressions and diverse movements across various style images, underscoring its robust performance.

\begin{table}[!t]
  \vskip -0.1in
  \caption{\textbf{Quantitative comparisons of emotion control .} The best results are in \textbf{bold}, and the second-best are in \underline{underlined}.}
  \vskip 0.1in
  \centering
  \scalebox{0.7}{
      \begin{tabular}{c|cccc}
      \toprule
      Methods & FID $\downarrow$ & FVD $\downarrow$ & LPIPS $\downarrow$ & Emo-A $\uparrow$ \\
      \hline
      EAMM \cite{ji2022eamm} & 111.710 & 210.275 & 0.223 & 0.160 \\
      DreamTalk \cite{ma2023dreamtalk} & 119.032 & 199.962 & 0.246 & 0.350 \\
      EDTalk \cite{tan2024edtalk} & 135.215 & 221.897 & 0.289 & \underline{0.460} \\
      EAT \cite{gan2023efficient} & \underline{95.085} & \underline{166.316} & \underline{0.138} & 0.450 \\
      Playmate(Ours) & \textbf{68.234} & \textbf{149.837} & \textbf{0.112} & \textbf{0.550} \\
      \bottomrule
      \end{tabular}
  }
  \label{tab:emo_control}
  \vskip -0.1in
\end{table}

\textbf{Emotion Control.} \cref{fig:emo_control} shows the generation results of our method under different emotional conditions using the same audio clip. These examples clearly illustrate our model's proficiency in encoding emotional signals into latent space and producing talking face animations with the desired emotion. This highlights Playmate's effectiveness in achieving precise emotional control while maintaining natural and lifelike results. Furthermore, we compare our method with several existing expression control approaches across various quantitative metrics on the MEAD test dataset. As presented in \cref{tab:emo_control}, Playmate achieves the highest scores in both video quality and emotion control metrics. 

\begin{table}[!h]
  \vskip -0.1in
  \caption{\textbf{User study comparison on open dataset.} The best results are in \textbf{bold}, and the second-best are in \underline{underlined}.}
  \vskip 0.1in
  \centering
  \scalebox{0.8}{
      \begin{tabular}{c|cccc}
      \toprule
      Methods & LS $\uparrow$ & VD $\uparrow$ & N $\uparrow$ & VA $\uparrow$ \\
      \hline
      JoyVASA \cite{cao2024joyvasa} & 2.500 & 2.286 & 1.714 & 1.929 \\
      Hallo \cite{xu2024hallo} & 2.964 & 2.929 & 3.071 & 2.893 \\
      Hallo2 \cite{cui2024hallo2} & 3.036 & 2.929 & 2.893 & 2.786 \\
      MEMO \cite{zheng2024memo} & 3.321 & 3.036 & 3.179 & 3.143 \\
      Sonic \cite{ji2024sonic} & \textbf{3.821} & \underline{3.071} & \textbf{3.750} & \underline{3.500} \\
      Playmate(Ours) & \underline{3.750} & \textbf{3.857} & \underline{3.464} & \textbf{3.643} \\
      \bottomrule
      \end{tabular}
  }
  \label{tab:user_study}
  \vskip -0.1in
\end{table}

\begin{figure}[t]
    \centering
    \includegraphics[width=0.9\linewidth]{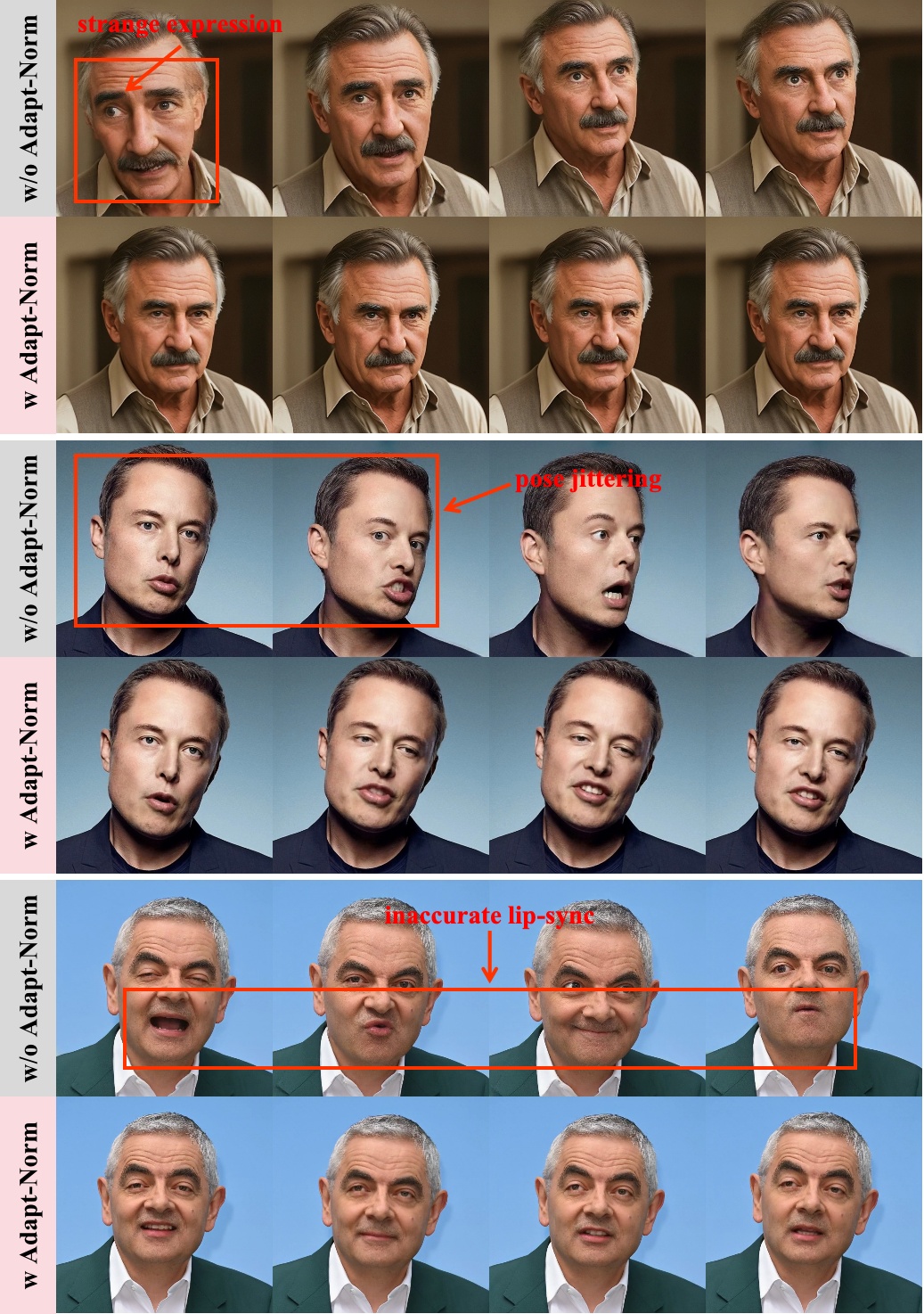}
    \caption{\textbf{Ablation study of Adaptive Normalization.} Without adaptive normalization, the generated results are prone to pose jittering, inaccurate lip-sync, and strange expressions, resulting in temporal discontinuities and artifacts within the final video.}
    \label{fig:ablation_adapt_norm}
    \vskip -0.1in
\end{figure}

\textbf{User Study.} We conducted a user study involving 10 participants who rated videos using the MOS (Mean Opinion Score) rating method, on a scale of 1 to 5, across four metrics: Lip Sync (LS), Video Definition (VD), Naturalness (N), and Visual Appeal (VA). As illustrated in \cref{tab:user_study}, Playmate has a notable advantage in the VD and VA metrics. While the LS and N metrics are slightly lower than Sonic's, they still outperform those of other methods, showcasing Playmate's strong competitiveness.

\subsection{Ablation Studies}
\textbf{Adaptive Normalization.} We analyze the effectiveness of our proposed Adaptive Normalization in decoupling expressions and head pose. As shown in \cref{fig:ablation_adapt_norm}, without adaptive normalization, the generated results are prone to pose jittering, inaccurate lip-sync, and strange expressions, resulting in temporal discontinuities and artifacts within the final video. After introducing adaptive normalization, where head pose feature and expression feature are normalized using different mean and variance values, the model can independently identify and distinguish pose feature and expression feature. Consequently, the final generated videos maintain better continuity in both pose and expression. 

\textbf{CFG scales.} By adjusting the CFG scales, we can strike a balance between the quality and diversity of the generated results. In \cref{tab:ablation_cfg}, we evaluate the selection of CFG scales for the audio and emotion conditions (represented by $w_a$ and $w_e$ in \cref{eq:cfg}) within our model.
Firstly, we conduct ablation experiments solely on $w_a$, as shown in the upper part of \cref{tab:ablation_cfg}. When $w_a$ is set to 1.5 or 2.0, Playmate achieves better evaluation metric scores. Upon comprehensive consideration, we set $w_a$ to 1.5 and conduct an additional ablation experiments on $w_e$. To more precisely assess the influence of $w_e$, we utilize the Emo-A metric, which helps in selecting a suitable value for $w_e$.
As shown in the bottom part of \cref{tab:ablation_cfg}, when $w_a$ is set to 1.5, setting $w_e$ to 1.5 as well can achieve more balanced results across various testing metrics. Therefore, we also set $w_e$ to 1.5.
\begin{table}[!t]
  \vskip -0.1in
  \caption{\textbf{Ablation study of the audio and emotion CFG scales.} The top half shows the ablation results for $w_a$, and the bottom half shows the ablation results when both $w_a$ and $w_e$ are combined. The best results are in \textbf{bold}, and the second-best are in \underline{underlined}.}
  \vskip 0.1in
  \centering
  \scalebox{0.6}{
      \begin{tabular}{cc|ccccccc}
      \toprule
      $w_a$ & $w_e$ & FID $\downarrow$ & FVD $\downarrow$ & Sync-C $\uparrow$ & Sync-D $\downarrow$ & CSIM $\uparrow$ & LPIPS $\downarrow$ & Emo-A $\uparrow$ \\
      \hline
      1.0 & $\emptyset$ & 17.128 & \textbf{118.481} & 7.937 & 7.22 & \textbf{0.939} & \textbf{0.044} & $\emptyset$ \\
      1.5 & $\emptyset$ & 17.096 & \underline{120.678} & \textbf{8.141} & \underline{7.064} & 0.935 & \underline{0.046} & $\emptyset$ \\
      2.0 & $\emptyset$ & \textbf{15.968} & 122.718 & \underline{8.113} & \textbf{7.049} & \underline{0.936} & \textbf{0.044} & $\emptyset$ \\
      2.5 & $\emptyset$ & \underline{16.887} & 125.738 & 8.03 & 7.109 & \underline{0.936} & 0.047 & $\emptyset$ \\
      \hline
      1.5 & 1.5 & 18.948 & \textbf{138.511} & \textbf{7.395} & \textbf{7.644} & \textbf{0.922} & \textbf{0.045} & 54.405 \\
      1.5 & 2.0 & 19.978 & 173.114 & \underline{7.205} & \underline{7.931} & \underline{0.905} & 0.056 & \textbf{57.579} \\
      1.5 & 2.5 & \textbf{17.498} & \underline{169.053} & 7.181 & 7.933 & 0.888 & \underline{0.054} & \underline{56.276} \\
      1.5 & 3.0 & \underline{17.825} & 177.528 & 7.16 & 8.075 & 0.888 & 0.058 & 50.055 \\
      1.5 & 3.5 & 22.32 & 207.535 & 6.89 & 8.313 & 0.871 & 0.067 & 55.753 \\
      \bottomrule
      \end{tabular}
  }
  \label{tab:ablation_cfg}
  \vskip -0.1in
\end{table}

\section{Conclusion}
In summary, we present Playmate, a two-stage training framework designed to generate lifelike talking videos guided by speech audio clips and various optional control conditions. Our approach addresses critical limitations of existing methods, such as precise motion decoupling, expression controllability, and lip synchronization accuracy. In the first stage, Playmate utilizes a motion-decoupled module to enhance attribute disentanglement accuracy and trains a diffusion transformer to produce expressive talking videos directly from audio cues. In the second stage, an emotion-control module is introduced to encode emotion control information into the latent space, enabling fine-grained control over emotions, thereby achieving the ability to generate talking videos with desired emotion. Extensive evaluations show that Playmate consistently surpasses existing state-of-the-art approaches in video quality and facial dynamics realism, while also demonstrating strong performance in lip synchronization and enhanced control over emotion and head pose. 

\textbf{Limitations and future work.} While Playmate shows significant advancements, it still has some limitations. Playmate processes information primarily around the face area. Extending its capability to the full upper body or even the whole body could provide additional capability. When using 3D implicit representations, the lack of a more explicit 3D face model may lead to artifacts such as blurred edges during extreme head movements, texture sticking due to neural rendering, and minor inconsistencies in complex backgrounds. Future work will focus on enhancing the model's robustness to diverse perspectives and styles by incorporating more diverse training data, as well as improving the rendering quality of the framework through advanced techniques.

\section*{Impact Statement}


This paper presents work whose goal is to advance the field of machine learning in portrait animation in particular. There are many potential societal consequences of our work, none of which we feel must be specifically highlighted here. For an extensive discussion of the general ramifications of talking video generation, we point interested readers towards \cite{xue2024human,jiang2024audio}.

\nocite{langley00}

\bibliography{Playmate_reference}
\bibliographystyle{icml2025}

\newpage
\appendix
\onecolumn
\section{Appendix}

In this appendix, we provide the following materials: 
\begin{enumerate}
  \item More visual comparisons of different methods. Refer to \cref{subsec:a_1} in the appendix.
  \item More details of Gaussian noise augmentation. Refer to \cref{subsec:a_2} in the appendix.
  \item Evaluate the accuracy of pose conditional control. Refer to \cref{subsec:a_3} in the appendix.
  \item Disentanglement of facial representation. Refer to \cref{subsec:a_4} in the appendix.
\end{enumerate}

\begin{figure*}[!b]
    \centering
    \includegraphics[width=0.9\linewidth]{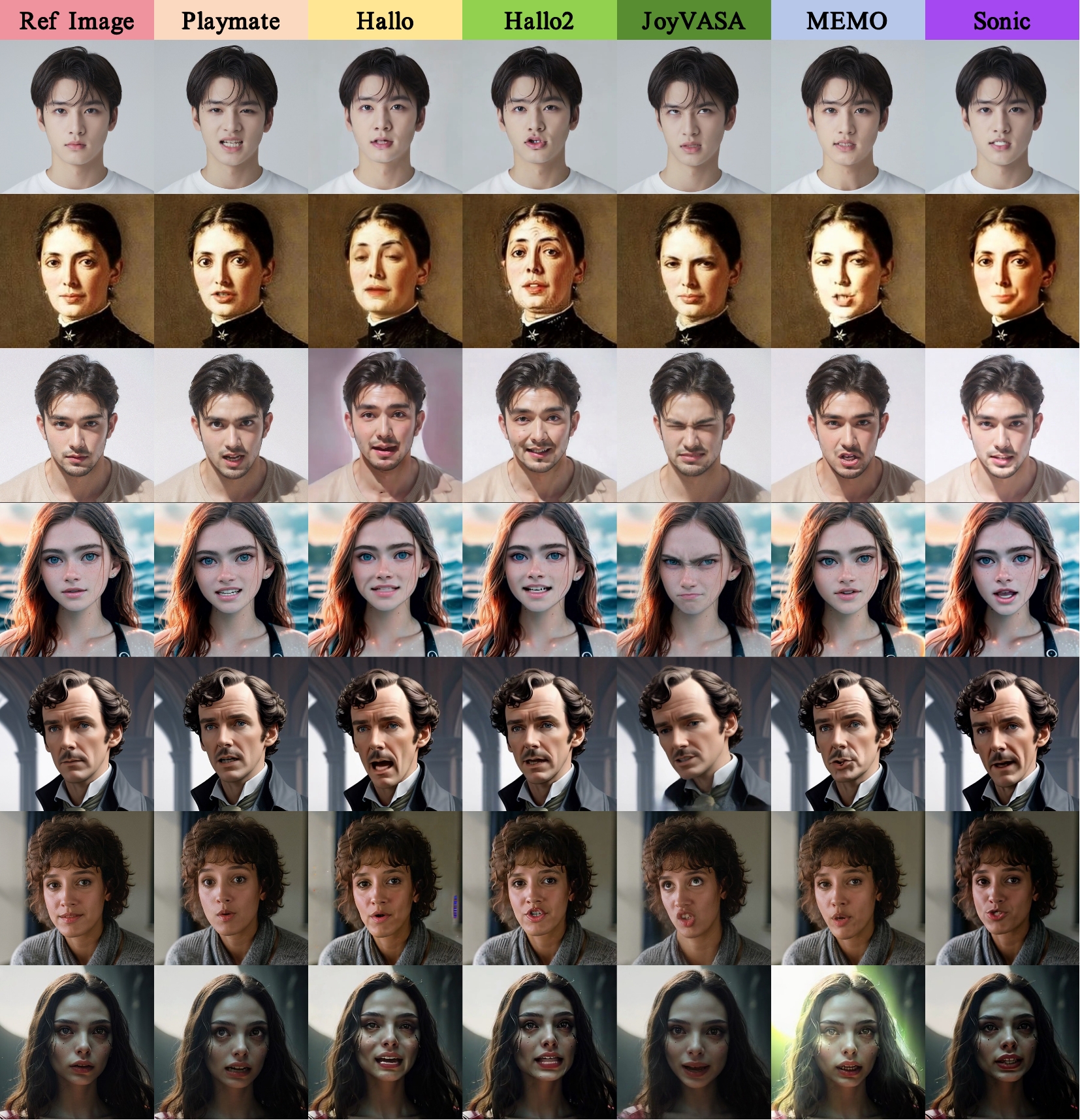}
    \caption{\textbf{More Visual Comparisons.} Each row of images represent the generation results of different methods using the same reference image and the same driving audio.}
    \label{fig:appendix_1}
\end{figure*}

\subsection{More Visual Comparisons}
\label{subsec:a_1}
To more comprehensively and objectively demonstrate Playmate's generation results, we compare the performance of Playmate and several other methods across a wider range of images. As shown in \cref{fig:appendix_1}, we present a comparison of the effectiveness of Playmate with other current methods in generating talking videos. As evident from the illustration, Hallo and Hallo2 exhibit noticeable artifacts in video quality, characterized by insufficient clarity in teeth generation and inaccurate lip-sync, as exemplified by the images in row 4, column 3 and row 2, column 4 respectively. JoyVASA, despite utilizing a 3D-Implicit Space approach for guidance, falls short in terms of pose and facial motion, resulting in notable head distortions, as exemplified by the images in row 1, column 5 and row 3, column 5. MEMO, on the other hand, struggles with exposure issues, often presenting artifacts on both the face and background, as exemplified by the images in row 2, column 6 and row 7, column 6. Finally, Sonic demonstrates good lip-sync and vivid expressions but still grapples with blurred teeth, as exemplified by the images in row 1, column 7 and row 4, column 7. In sharp contrast, our proposed method showcases the ability to generate more natural facial expressions and head movements that are well-synchronized with audio inputs. Additionally, videos produced by Playmate exhibit superior overall visual quality and stronger identity consistency.

\subsection{Gaussian Noise Augmentation}
\label{subsec:a_2}
\begin{figure*}[!b]
    \centering
    \includegraphics[width=0.8\linewidth]{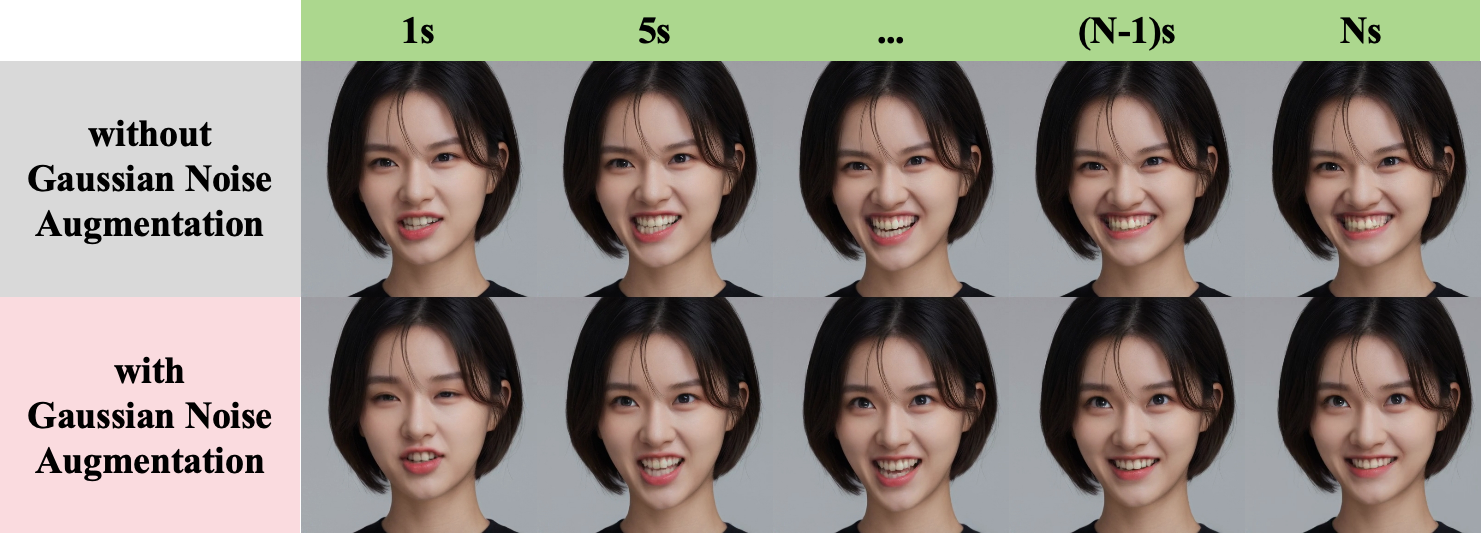}
    \caption{Ablation study of the Gaussian noise augmentation.}
    \label{fig:appendix_2}
\end{figure*}
We additionally utilize the speech and motion representation of the previous video frames as input to the audio conditioned diffusion transformer, aiming to maintain the continuity of the generated video. But in this incremental generation process, some contaminations of the previously generated video frames, such as the background noise and subtle distortions in facial expressions, will propagate to subsequent frames and continue to amplify artifacts. In order to enhance the resistance of our diffusion transformer to the above contaminations, we similar to Hallo2 incorporate Gaussian noise into the prev motion representations as: 
\begin{equation}
\begin{split}
\hat{m}^{prev}= \sqrt{\bar{\alpha}_t} \tilde{m}^{prev}+ (1 - \bar{\alpha}_t) \epsilon ,\epsilon \sim \mathcal{N}(0, \mathbf{I}), t \in (0, 50),
\end{split}
\end{equation}
where $\tilde{m}^{prev}$ represents the motion representations of the previous frame, and $\epsilon$ represents the Gaussian noise of the corresponding frame. The Gaussian noise follows the normal distribution $\mathcal{N}(0, \mathbf{I})$, in which $\mathbf{I}$ denotes the identity matrix. These noise-augmented motion representations are concatenated with current noisy motions $m_t$ and jointly participate in the diffusion process. Specifically, each denoising step can be described as:
\begin{equation}
\begin{split}
\hat{\epsilon}_{\theta}\left(m_{t}, f_a,f_{id}, t\right)=\hat{\epsilon}_\theta \left(concat(\hat{m}^{prev}, m_t),f_a,f_{id},t\right),
\end{split}
\end{equation}
where $\hat{\epsilon}_{\theta}\left(m_{t}, f_a,f_{id}, t\right)$ represents the noise component predicted by Audio Conditioned Diffusion Transformer, and $f$ represents the conditioning inputs of audio feature and identity feature. Through this noise-augmented operation, our diffusion transformer is more robust to slight changes in the motion input, thereby mitigating the impact of contaminations from previously generated video frames. 

We conducted an ablation study on Gaussian noise augmentation, as illustrated in \cref{fig:appendix_2}. Without this augmentation strategy, once artifacts emerged in the generated image, these flaws would propagate, leading to flawed generations in subsequent images. After applying the augmentation strategy, we observed a significant improvement in the quality of the generated images.

\subsection{Evaluation of Pose Control}
\label{subsec:a_3}
In the field of image animation, the mean keypoint distance (AKD) and the average pose distance (APD) are commonly used to evaluate the performance of pose control. We calculated the APD metric of Playmate on two datasets (HDTF and our dataset), as shown in the table below.
\begin{table*}[h]
  \vskip -0.1in
  \caption{Evaluation of Pose Control.}
  \vskip 0.1in
  \centering
  \scalebox{0.9}{
  \begin{tabular}{c|ccc}
  \toprule
  Dataset & $APD_{jaw}$ & $APD_{pitch}$ & $APD_{roll}$ \\\hline
  HDTF & $3.003^\circ$ & $1.308^\circ$ & $1.214^\circ$ \\
  Collected dataset & $3.714^\circ$ & $1.751^\circ$ & $1.398^\circ$ \\
  \bottomrule
  \end{tabular}%
  }
  \label{tab:eva_pose_control}%
\end{table*}

\subsection{Disentanglement of Facial Representation}
\label{subsec:a_4}
\begin{figure*}[h]
    \centering
    \includegraphics[width=0.9\linewidth]{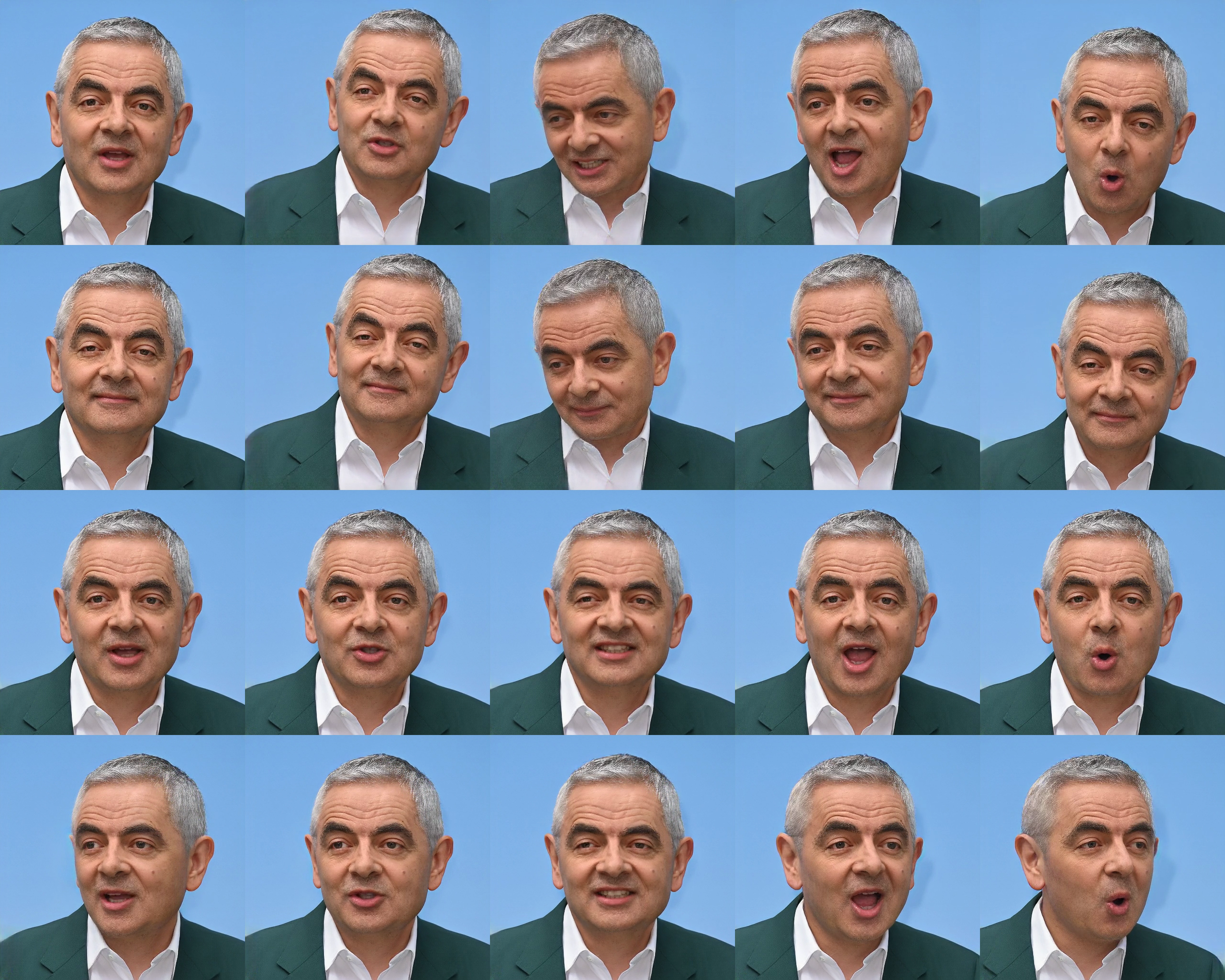}
    \caption{Disentanglement between head pose and facial dynamics.}
    \label{fig:appendix_4}
\end{figure*}

\cref{fig:appendix_4} shows the effective disentanglement of the latents of the face. From top to bottom: the raw generated sequence, applying generated poses with fixed initial facial dynamics, and applying generated facial dynamics with fixed initial head pose and predefined spinning poses.

\end{document}